%% file: main.tex
\documentclass[journal]{IEEEtran}
\ifCLASSINFOpdf
\else
\fi

\input{micro}
\begin{document}

\title{NARU: A Benchmark for NARrative Evolution and Cultural Nuance Understanding in Japanese Extreme Long Video}

\author{Yuheng~Huang*,
        Jianlang~Chen*,
        Jiayang~Song,
        Hua~Qi,
        Aza~Kai,
        Vincent~Markert,
        Edison~Marrese-Taylor,
        Jianjun Zhao
        and~Lei~Ma,~\IEEEmembership{Member,~IEEE}
\thanks{Yuheng Huang and Jianlang Chen contributed equally to this work.}
\thanks{Yuheng Huang, Hua Qi and Lei Ma are with The University of Tokyo, Tokyo, Japan (e-mail: yuhenghuang42@g.ecc.u-tokyo.ac.jp; qi-hua@g.ecc.u-tokyo.ac.jp; ma.lei@acm.org).}%
\thanks{Jianlang Chen is with Kyushu University, Fukuoka, Japan (e-mail: chen.jianlang.396@s.kyushu-u.ac.jp).}%
\thanks{Jiayang Song is with Macau University of Science and Technology, Macau, China (e-mail: jiayang.song@ieee.org).}%
\thanks{Aza Kai, Vincent Markert, and Edison Marrese-Taylor are with Infinimind Japan Inc., Japan (e-mail: kai@infinimind.io; vincent@infinimind.io; edison@infinimind.io). Edison Marrese-Taylor is also with The University of Tokyo, Japan.}%
\thanks{Jianjun Zhao is with Kyushu University, Fukuoka, Japan (e-mail: zhao@ait.kyushu-u.ac.jp)}
\thanks{Lei Ma is also with University of Alberta, Edmonton, Canada.}%
}


\newcommand{\EMT}[1]{\textcolor{green}{{\small [#1 --Edison]}}}

\markboth{Journal of \LaTeX\ Class Files,~Vol.~14, No.~8, August~2026}%
{Shell \MakeLowercase{\textit{et al.}}: Bare Demo of IEEEtran.cls for IEEE Journals}

\maketitle
\begin{abstract}
Long-form video understanding encompasses tasks that go beyond retrieving isolated events, including tracking an evolving narrative and interpreting social meaning that may remain implicit. However, existing benchmarks rarely evaluate these capabilities jointly, particularly in high-context, non-English media. To address this gap, we introduce {\ourbench}, a benchmark designed to evaluate Narrative evolution and Reasoning on cultural Understanding in Japanese long-form video. {\ourbench} consists of 1,481 questions grounded in 155 videos totalling 146.8 hours, spanning four narrative and five cultural dimensions. To construct the benchmark at this scale, we propose a hierarchical memory-based annotation pipeline that transforms raw video into structured event, narrative, and cultural annotations, then generates questions via task-oriented synthesis and iterative shortcut removal. The construction process includes two native-speaker verification stages involving 68 annotators.  Evaluations across eight model configurations reveal substantial limitations in both long-range narrative integration and culturally grounded reasoning. By exposing these persistent gaps, {\ourbench} offers a systematic testing ground for developing MLLMs capable of reliably interpreting long-form, high-context video.
\end{abstract}

\section{Introduction}

Recent advances in Multimodal Large Language Models (MLLMs) have substantially improved their abilities in video understanding, supporting accurate captioning and visual question answering for short video clips and temporal events~\cite{wang2025internvl3, bai2025qwen2, zhang2025videollama3}.
These capabilities have further enabled a growing range of new downstream applications, such as short-form content retrieval, clip-level summarization, and open-domain video commentary generation~\cite{tang2025video,marrese-taylor-etal-2022-open}. 
However, in addition to short video clips, many real-world scenarios involve long-form video content~\cite{zou2024seconds}, such as full-length films and television episodes, livestream archives, and documentaries, where users may seek a holistic understanding across the entire video rather than isolated facts from individual fragments. 
Tasks such as long-form video summarization~\cite{hua2025v2xum}, content analysis~\cite{song2024moviechat, faure2025hermes}, culturally aware content moderation~\cite{mukherjee2025toward, chen2025memearena}, and media archiving~\cite{kriz2025multivent} require models to integrate information across extended time spans and reason over evolving social and narrative contexts.




Crucially, a video's length alone does not dictate the need for global context.
For example, hours-long continuous surveillance footage may support tasks that can be solved through localized detection or retrieval. In contrast, films, television programs, documentaries, and other socially situated media often contain events whose significance emerges only through their relationships to earlier events, evolving participants, and accumulated social context. We refer to such content as \emph{context-rich long-form video}. Understanding such content imposes two coupled demands. First, models need to preserve narrative coherence by tracking entities, events, causal dependencies, and character development across the video. Second, they are required to interpret brief but consequential moments whose significance depends on accumulated social context rather than literal audiovisual content alone.

Existing benchmarks have advanced the evaluation of these capabilities from various aspects. LongVideoBench~\cite{wu2024longvideobench} and LVBench~\cite{wang2025lvbench} evaluate long-range retrieval, temporal grounding, and general reasoning over extended videos, while recent narrative-oriented benchmarks such as VRBench~\cite{yu2025vrbench} and StoryVideoQA~\cite{wu2026storyvideoqa} target multi-step reasoning and deep storyline comprehension. 
Multilingual benchmarks such as ViMUL-Bench~\cite{shafique2025culturally} further broaden video evaluation across languages and cultural categories. Nevertheless, these directions remain largely separate: existing benchmarks do not jointly examine sustained narrative comprehension and culturally situated implicit reasoning in context-rich long-form video.



\begin{figure}[ht]
    \centering

    \begin{subfigure}[t]{\linewidth}
        \centering
        \includegraphics[width=0.95\linewidth]{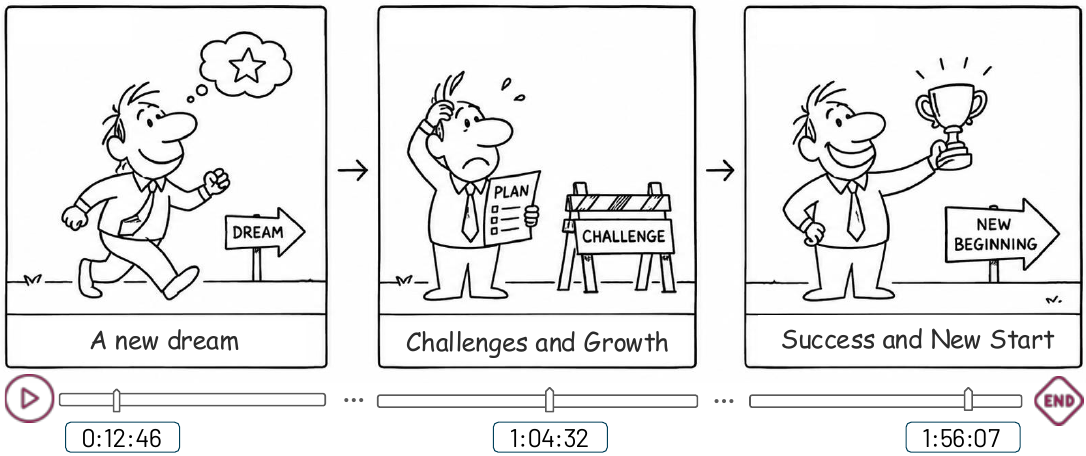}
        \caption{Narrative Intelligence: The model is asked to summarize a character's storyline over a long video.}
        \label{fig:demo:narrative}
    \end{subfigure}

    \vspace{0.5em}

    \begin{subfigure}[t]{\linewidth}
        \centering
        \includegraphics[width=0.95\linewidth]{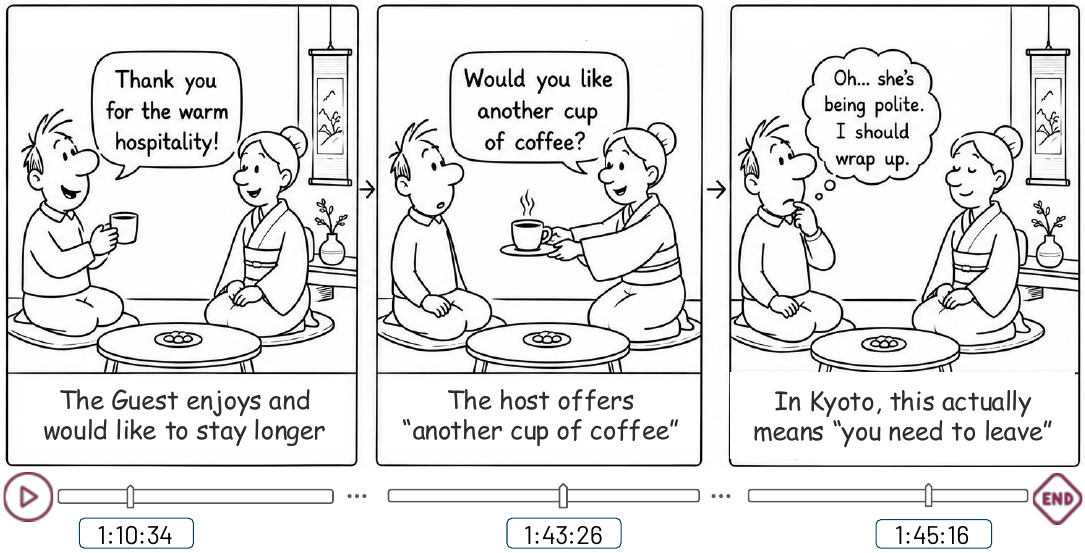}
        \caption{Cultural Understanding: The model is asked to infer the host's implicit intent when offering another cup of coffee to the guest.\footnotemark}
        \label{fig:demo:culture}
    \end{subfigure}
    \vspace{-10pt}
    \caption{Example scenarios from {\ourbench}.}
    \label{fig:demo}
    \vspace{-10pt}
\end{figure}

Japanese long-form media is both a practically consequential domain and a scientifically informative setting for context-rich video understanding. Japan supports a substantial video-content ecosystem, with its domestic market projected to reach approximately 630 billion yen in fiscal year 2025~\cite{Yano2025VideoContent}. This scale motivates evaluating whether MLLMs can understand Japanese media. More importantly, Japanese communication is characterized by a high-context cultural style, in which meaning is often conveyed implicitly through ambient atmosphere (kuuki wo yomu, 空気を読む, or ``reading the air''), conversational backchannels (\emph{aizuchi}, 相槌), and culturally shared expectations. 
Understanding these interactions requires models to integrate diffuse cues across extended temporal spans and to maintain a coherent interpretation of evolving interpersonal and narrative dynamics. {\ourbench} therefore uses Japanese long-form media as a focused setting for evaluating context-rich video understanding, rather than treating temporal duration, narrative reasoning, and cultural knowledge as independent capabilities. 


\footnotetext{In certain situations in Japan, such offers are conventionally used as an indirect and polite way to signal the end of a visit.
Understanding the host's true intent requires awareness of the context and the surrounding social atmosphere, beyond the literal semantic content.}

To address these gaps, we introduce {\ourbench} (NARrative and Cultural Understanding), a multimodal benchmark designed to evaluate MLLMs on extreme long-form video understanding within a high-context cultural framework. {\ourbench} focuses on two core competencies, as demonstrated in Fig.~\ref{fig:demo}:
(1) Narrative Intelligence, which measures a model's ability to track story evolution across long videos with a duration of 30 to 240 minutes, and
(2) Native Cultural Understanding, which assesses comprehension of Japanese conversational nuances, social dynamics, and implicit communicative signals.

Constructing a benchmark of this scale exposes fundamental limitations in traditional dataset creation paradigms. Existing high-quality benchmarks rely heavily on manual annotation pipelines, which are effective for short clips but become unscalable and cognitively unreliable when applied to extremely long videos in culturally specific domains.

To address these challenges, we introduce a hierarchical annotation-to-QA synthesis pipeline for long-form video. The pipeline decomposes videos into temporal segments, maintains cross-segment narrative continuity, and supports multi-level annotations ranging from fine-grained events to high-level narrative and social analysis. This approach enables scalable, context-aware ground truth construction that is infeasible with conventional manual annotation alone. Using this pipeline, we build a high-quality long-form video understanding benchmark, followed by expert verification to ensure annotation reliability and consistency.
In summary, this paper makes the following contributions:

\begin{itemize}[leftmargin=*]
\setlength\itemsep{0.3mm}
    \item \textbf{Problem Formulation}: We distinguish context-rich long-form video from video that is merely long in duration and formulate its understanding as the joint problem of maintaining narrative state and interpreting culturally situated implicit meaning.

    \item \textbf{Benchmark Development}: We introduce {\ourbench}, a large-scale benchmark comprising 155 Japanese long-form videos totaling 146.8 hours and 1,481 multiple-choice questions, designed to systematically evaluate Narrative Intelligence and Native Cultural Understanding in extreme long-form settings. The final items are verified by a total of 68 native Japanese annotators to ensure video grounding and cultural fidelity.

    \item \textbf{Technical Contribution}: We introduce a hierarchical annotation-to-QA synthesis pipeline that constructs temporally coherent annotations for hours-long videos and uses them to generate context-dependent question–answer pairs.

    \item \textbf{Systematic Evaluation}: We benchmark SOTA MLLMs on {\ourbench}, establishing baselines for context-rich Japanese long-form video understanding across narrative and culturally implicit reasoning dimensions.
\end{itemize}

The benchmark data, evaluation resources, and additional details are available on our project website~\cite{ourwebsite}.

\section{Related Work}

Video question-answering benchmarks initially focused on short clips, evaluating local event recognition, action understanding, and temporal reasoning~\cite{xu2017video,xiao2021next,maaz2024video,li2024mvbench}. Recent benchmarks extend evaluation to movies, television programs, documentaries, and other long-form videos~\cite{song2024moviechat,zhou2025mlvu,rawal2024cinepile,fu2025video,wu2024longvideobench,wang2025lvbench,ma2026scalelong}. These efforts substantially broaden temporal coverage, evaluating capabilities such as evidence retrieval, temporal grounding, event understanding, summarization, and multi-timescale integration. However, extended temporal coverage alone does not capture whether a model can maintain an evolving interpretation of narrative and social context.

To bridge this gap, story-focused benchmarks examine complementary aspects of narrative comprehension. DramaQA introduced hierarchical, character-centered question answering, while SCVBench evaluates story-centric temporal reasoning~\cite{choi2021dramaqa,you2025scvbench}. More recent benchmarks further address multi-step reasoning, entity persistence, distributed evidence, and long-range storyline comprehension~\cite{yu2025vrbench,ha2026narrativetrack,jain2026navqa,wu2026storyvideoqa}. While these efforts successfully move beyond localized event recognition, they primarily formulate narrative understanding through structural plot elements, entity continuity, temporal relations, and explicit causal dependencies.

In parallel, social and cultural reasoning present another crucial layer of video understanding. Prior work has explored this through multimodal social reasoning, geographically diverse visual knowledge, and multilingual video evaluation~\cite{zadeh2019socialiq,guo2023desiq,nayak2024culturalvqa,shafique2025culturally}. Although these benchmarks provide vital coverage of social interaction and cultural knowledge, they rarely examine how culturally implicit meanings emerge and evolve across an extended narrative arc.
{\ourbench} connects these research directions through \emph{context-rich long-form video understanding}. Its narrative dimension requires models to track characters, events, causal relations, and plot development across extended temporal horizons, while its cultural dimension requires them to infer socially implicit meanings from evolving interpersonal context.

\begin{figure*}[t]
    \centering
    \includegraphics[width=\textwidth]{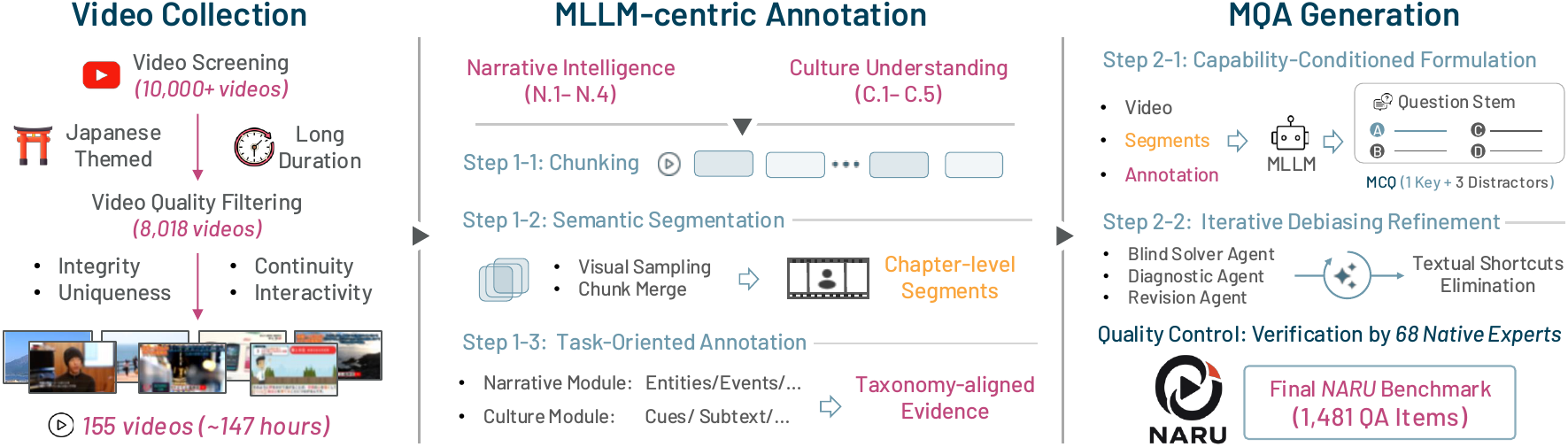}
    \caption{Workflow overview of {\ourbench}. Video collection and filtering (Sec.~\ref{sec:method:video}) select 155 long videos from the candidate set. A MLLM-centric pipeline (Sec.~\ref{sec:method:annotation}) is used to produce taxonomy-aligned evidence for narrative intelligence and cultural understanding annotation. 
    Multiple-choice questions (Sec.~\ref{sec:method:Generation}) are subsequently generated and refined through a multi-agent pipeline. Verification (Sec.~\ref{sec:method:quality}) by 68 native Japanese experts yields the final 1,481 QA items.}
    \label{fig:overall_workflow}
    \vspace{-10pt}
\end{figure*}

\section{NARU Benchmark Construction}

While Japanese long videos can serve as an ideal testbed for evaluating MLLMs' abilities, we face a dual dilemma in constructing a high-quality benchmark from them. 
On one hand, relying solely on manual annotations is impractically costly and unscalable.
Namely, annotating extremely long videos requires native Japanese experts to watch hours of content, track complex narrative arcs, and identify subtle cultural cues. 
This process can incur high cognitive load and time costs. 
Additionally, while automated MLLM pipelines offer scalability, they are currently hindered by the context-window bottleneck. 
Even SOTA models struggle to process hours of continuous video without losing fidelity, often hallucinating details or failing to maintain narrative consistency over long durations~\cite{wang2025lvbench}.

To bridge the gap, we propose a hierarchical memory-based annotation pipeline. The system performs short-term chunk processing with long-term narrative history to analyze and annotate multimodal information across extended timelines. 
This pipeline provides experts with abundant contextual information, substantially reducing routine annotation effort and allowing them to focus on identifying and resolving more complex, high-impact cases. 
We describe the construction process of {\ourbench} through Sec.~\ref{sec:method:taxonomy} to Sec.~\ref{sec:method:quality}.

\subsection{Capability Taxonomy}
\label{sec:method:taxonomy}

To evaluate long-form video understanding beyond mere information retrieval, we organize {\ourbench} around two complementary capabilities. First, \textbf{narrative intelligence} measures a model's ability to track and synthesize evolving events, entities, and concepts across the extended runtime. Second, \textbf{cultural understanding} evaluates how well a model interprets implicit, socially situated meanings that transcend literal audiovisual signals. Rather than relying on a single existing framework, we bridge theories of event cognition and high-context communication to construct subcategories with distinct inference targets.

\noindent\textbf{Narrative Intelligence (N)}

To ground our evaluation of narrative structure, we draw on Event Segmentation Theory (EST)~\cite{zacks2007event}. EST posits that humans perceive continuous experience as a structured sequence of discrete events, maintaining internal event models that are updated at situational boundaries. 
Therefore, we distinguish between \textbf{local coherence}, which captures consistency across adjacent or tightly linked events, and \textbf{global coherence}, which requires integrating evidence distributed throughout the video into a unified interpretation. We operationalize these two levels of narrative structure through four subcategories:

\begin{itemize}[leftmargin=*]

\item \textit{N.1: Character/Entity Evolution} (Local Coherence): The ability to track a recurring character or entity across temporally separated segments and infer how its state evolves over time. Relevant states include goals, beliefs, emotions, roles, relationships, and circumstances. This category evaluates whether a model can maintain a consistent representation of an entity while continuously updating it in response to new events, rather than treating each appearance independently.

\item \textit{N.2: Sequential/Topical Flow} (Local Coherence): The ability to reconstruct how a video progresses from one event, scene, or topic to the next. This includes recognizing temporal order, identifying transitions and topic shifts, and relating each segment to its immediate context. Different from N.1, which centers on entity evolution, N.2 evaluates the continuity and organization of the narrative itself.

\item \textit{N.3: Plot/Conflict Progression} (Global Coherence): The ability to trace the evolution of goals, obstacles, conflicts, and consequential decisions across multiple events. This includes identifying initiating conditions, major turning points, causal consequences, and eventual resolutions or shifts in stakes. Rather than simply recovering temporal order, this category evaluates whether a model can integrate dispersed events into a coherent account of how the central narrative unfolds.

\item \textit{N.4: Idea/Thematic Development} (Global Coherence): The ability to synthesize the higher-level ideas, arguments, values, or themes developed throughout a video and explain how they are introduced, supported, refined, or reinterpreted over time. Unlike N.3, which focuses on narrative progression, this category captures forms of global coherence organized around conceptual or thematic development, particularly in documentaries, interviews, and discussion-oriented content.
\end{itemize}

Together, N.1 and N.2 assess whether a model can maintain a stable account of \emph{who or what is involved} and \emph{how the content proceeds}. 
N.3 and N.4 test whether it can infer \emph{why the events matter} within the narrative and \emph{what broader meaning they jointly express}.




\noindent\textbf{Cultural Understanding (C)}

For cultural understanding, we focus on interactions in which meaning depends on shared social assumptions and contextual cues beyond explicit verbal content~\cite{hall1976beyondculture}. We organize the taxonomy according to an evidence-to-interpretation process. C.1 evaluates the interpretation of an observable conversational signal. C.2, C.3, and C.5 evaluate latent social meaning at three distinct levels: the shared situation, an individual speaker's intent, and the participants' affective states, respectively. C.4 focuses on the culturally specific background knowledge that can support interpretation at any of these levels.

\begin{itemize}[leftmargin=*]
    \item \textit{C.1: Aizuchi} (Interactional Signalling): 
    \textit{Aizuchi} are short listener utterances in Japanese conversation that signal attention, understanding, or engagement without necessarily indicating agreement (roughly like the English expressions “uh-huh” and “yep”)~\cite{kita2007nodding}.
    This category assesses a model's ability to interpret Japanese backchannels using their lexical, prosodic, temporal, and interactional context. 
    The same backchannel can indicate agreement, surprise, emotional alignment, or an attempt to manage turn-taking~\cite{maynard1986backchannel}. 
    We aim to understand whether a model can recover the conversational function of such phatic expressions, rather than equating its surface form with a fixed meaning.

    \item \textit{C.2: Kuuki wo Yomu} (Shared Situational Understanding):
    \textit{Kuuki wo Yomu}, literally “read the air,” refers to inferring the unspoken social atmosphere and interactional norms from contextual and interpersonal cues, including group consensus, interpersonal tension, and expectations about appropriate behavior, from participants' actions. 
    This category evaluates whether a model can construct a situation-level account of what the group collectively recognizes and how that understanding affects participants' behavior accordingly.

    \item \textit{C.3: Subtext Interpretation} (Speaker Intent): The ability to infer what a particular speaker intends to communicate beyond, or in contrast to, the literal meaning of an utterance. This includes indirect requests, euphemism, irony, socially restrained expression, and distinctions between \emph{tatemae} (public-facing expression) and \emph{honne} (private intent). Different from C.2, which concerns the shared atmosphere or norms of the overall situation, C.3 targets the latent communicative intention behind a specific speaker's expression.

    \item \textit{C.4: Cultural Context Recognition} (Cultural Grounding): The ability to recognize culturally specific objects, practices, social conventions, and historical references and to explain their significance within the current scene. This category evaluates the background knowledge needed to understand the specific meaning behind an observed action or reference, rather than merely checking if the model can visually identify it.

    \item \textit{C.5: Sentiment Analysis} (Affective and Interpersonal Dynamics): The ability of leveraging evidence to infer participants' affective states and interpersonal attitudes, and to track how they evolve throughout an interaction. Evidence may include speech content, facial expressions, and body language. This category encompasses social norms, such as concealed discomfort, restrained frustration, or emerging affinity, rather than only explicit emotional expressions or coarse positive–negative sentiment. Different from C.2 and C.3, this category evaluates how participants feel and relate to one another.
\end{itemize}

\subsection{Dataset Construction}
\label{sec:method:video}


\textbf{Acquisition and Pre-filtering.} 
To build a broad candidate pool, we searched across 16 YouTube-defined upload categories, continuing retrieval until reaching over 100,000 unique videos after deduplication. We then applied an ASR-based spoken-language identification pipeline, filtering for Japanese-language content, which yielded 51,643 videos. Finally, following the long-video criterion established by LVBench~\cite{wang2025lvbench}, we retained videos with durations of at least 30 minutes. This resulted in a final set of 8,018 candidate videos for subsequent content screening and diversity-aware selection.

\textbf{Video Quality Filtering.} A key challenge in evaluating long-context multimodal models is ensuring that long duration reflects meaningful progression rather than redundancy (\eg, videos with looping contents or disordered clip compilations).
Specifically, we required each video to exhibit temporally ordered changes in at least one benchmark-relevant dimension, such as character or entity states, events or actions, scenes or activities, topics or arguments, or interpersonal dynamics. We excluded repetitive or looping content, such as Rainstorm Sounds for Relaxing, in which different temporal portions are largely interchangeable with no evolving semantic information.
To translate this requirement into a concrete video selection process and ensure alignment with the capabilities defined in our taxonomy, two authors proficient in Japanese manually screened the candidate videos against four criteria:

\begin{enumerate}[leftmargin=*]
    \item \textit{Visual Integrity:} We excluded videos dominated by static imagery or obstructive watermarks because they provide insufficient evolving visual evidence for long-video understanding.
    \item \textit{Temporal Semantic Progression:} The reviewers inspected the content at the 25\%, 50\%, 75\%, and 100\% timestamps and scrubbed the timeline around each anchor point. At each location, they identified the active characters or entities, ongoing events or activities, and current topic or argument. A video was retained only when these observations exhibited temporally ordered changes in at least one dimension and the changes formed a connected sequence of events, topics, or arguments. We excluded videos whose sampled portions were repetitive or interchangeable, as well as compilations of unrelated clips.
\end{enumerate}

\textbf{Incremental Semantic Diversity Sampling.} To prevent over-representation of certain domains and maintain broad coverage of cultural contexts, we applied an automated sampling strategy guided by semantic diversity. 
Starting from 30 selected seed videos, we used a greedy, embedding-based selection process. A new video was added only if the cosine similarity between its title and description and those of all previously selected samples was below a fixed threshold $\tau$. 
This procedure ensures that each video contributes new semantic content, expanding the range of scenarios and perspectives represented in {\ourbench}. The full algorithm is described on our website~\cite{ourwebsite}. Finally, we selected 155 videos from the candidate set, corresponding to approximately 146.8 hours of content.

\subsection{MLLM-Driven QA Generation}

Given the selected videos and capability taxonomy, we construct candidate question--answer pairs through a \emph{hierarchical annotation-to-QA synthesis} pipeline. Direct generation from an hours-long video would require a single model invocation to preserve fine-grained evidence, integrate narrative relations, and interpret culturally implicit cues. We therefore separate benchmark construction into two stages. First, the \emph{\textbf{annotation stage}} transforms each video into a temporally connected hierarchy of chunk- and segment-level evidence and enriches this representation with taxonomy-aligned narrative and cultural annotations. Second, the \emph{\textbf{question--answer pair generation stage}} uses the resulting annotations to synthesize multiple-choice questions and reduce text-only shortcuts through video-blind diagnosis and targeted revision.

\subsubsection{\textbf{Annotation}}
\label{sec:method:annotation}

This process converts a long-form video into a structured representation using an MLLM-centric, multi-layer memory pipeline based on chunking, segmentation, and task-oriented annotation.

\textbf{Step 1-1: Chunking.}
Following prior long-video annotation protocols that use five-minute clips as units for dense narration~\cite{grauman2022ego4d,yang2025egolife}, we partition each video into approximately five-minute chunks. We treat this duration as a practical local processing unit rather than a semantic boundary. When transcripts are available, boundaries are aligned to the nearest transcript endpoint after five minutes to avoid splitting dialogue; otherwise, fixed-duration cuts are used. 
An MLLM processes these chunks sequentially to produce schema-constrained JSON records containing chunk summaries, entity lists, timestamped events, and closed captions (including dialogue, on-screen text, non-speech audio, and vocal tone). 
To maintain entity consistency and track cross-boundary events, the model is given a textual recap of preceding chunks with relative timestamps whenever it generates a new record. 
Finally, chunk records are unified into a global timeline by shifting timestamps to video-relative coordinates, deduplicating entities by identifier, and sorting events chronologically.


{\textbf{Step 1-2: Semantic Segmentation.}} 
Because chunk boundaries reflect processing constraints rather than narrative structure, we perform a second pass over the full video using lower-rate visual sampling and the merged Step~1-1 annotation as reference. Motivated by Segmented Discourse Representation Theory~\cite{lascarides2007segmented}, this pass identifies contiguous, chapter-level segments with a coherent topic or narrative function. Each segment records its temporal span, summary, detailed description, and content type. A boundary is introduced at a substantive thematic shift or a change in content function, such as a transition between the main program, ignoring routine camera cuts and speaker turns. The resulting segments are combined with the video-level entities, events, and closed captions to provide the structured context for the taxonomy-aligned annotation in Step~1-3.


{\textbf{Step 1-3: Task-Oriented Annotation.}}
Steps~1--2 record what occurs in the video and organize this evidence over time, but they do not explicitly capture the higher-level narrative and cultural relations defined in our benchmark taxonomy. We therefore introduce two complementary MLLM-based annotation modules \wrt to the narrative intelligence and cultural understanding capabilities mentioned in Sec.~\ref{sec:method:taxonomy}. 
The \emph{narrative annotation module} reasons over the video-level representation to assign functional roles to characters and organize events into coherent narrative threads. 
Each thread identifies its central entities, highlights consequential events, and summarizes their causal progression, providing structured evidence for questions on four subjects in narrative intelligence (N.1-N.4). 
The \emph{cultural annotation module} analyzes each semantic segment and extracts timestamped evidence for the five cultural dimensions (C.1-C.5). 
These segment-level annotations are subsequently integrated along the video timeline. Together, the two modules transform a general video representation into taxonomy-aligned evidence for subsequent question generation. 

\subsubsection{\textbf{Question-Answer Pair Generation}}
\label{sec:method:Generation}

Using the taxonomy-aligned annotations, we employ an MLLM to generate candidate four-option multiple-choice questions (MCQs). 
We organize this process into two steps: 
(1) capability-conditioned formulation, which associates each question with a taxonomy category and supporting evidence; 
(2) video-blind diagnosis and targeted revision, which identifies and repairs text-only shortcuts. All generated items remain candidates until the human verification described in Sec.~\ref{sec:method:quality}.

\textbf{Step 2-1: Capability-Conditioned Formulation.} The entry point for generation is the selection of a cognitive task grounded in the annotations provided in Sec.~\ref{sec:method:annotation}. 
For each evaluation category, the MLLM receives the original video, the segment-level representation, the relevant narrative or cultural annotation, and task-specific instructions and examples. 
Then, it is required to provide a question stem, a proposed correct answer option and a supporting record identifying the question rationale and its relevant entities, events, temporal segments, and audio cues. Simultaneously, the model generates three candidate distractor options. 
Distractor design is critical since implausible options or differences in length, specificity, emotional tone, and abstraction can reveal the correct answer without requiring the video (text-only shortcut). 
We therefore require all options in each MCQ to have comparable granularity, polarity, length, and syntactic complexity. 
Meanwhile, each distractor must also be plausible within the video's setting while remaining incorrect \wrt its evidence. 
These constraints provide an initial safeguard against option-level shortcuts before the video-blind screening in Step~2-2.



\textbf{Step 2-2: Iterative Debiasing Refinement.}
\label{sec:method:refinement}
Recent work shows that MLLMs can answer visual questions confidently without visual input, creating an illusion of grounded understanding~\cite{asadi2026mirage}. 
To alleviate such textual and language-bias shortcuts, we employ a Solver-Critic Loop, a three-stage iterative debiasing pipeline designed to empirically validate and refine question difficulty:

\begin{itemize}[leftmargin=*]
    \item \textbf{The Blind Solver Agent:} An agent attempts to answer the generated question without the video context ($V = \emptyset$). If this ``blind'' agent correctly identifies the answer, the question is flagged as potentially leaked.

    \item \textbf{Diagnostic Agent:} A reasoning agent analyzes the Blind Solver Agent's success to identify the vulnerability, such as \emph{Tone Bias} (\eg, the correct answer is the only polite option) or \emph{Process of Elimination}(\eg, several distractors contain implausible or mutually inconsistent details, leaving one viable answer without access to the video). 
    It then outputs a \emph{Refine Plan} to explain how such a shortcut happens and provides suggestions to address the underlying vulnerability.

    \item \textbf{Question Revision Agent:} Guided by the Refine Plan, this agent rewrites the question stem, correct answer, or distractors to eliminate language shortcuts.
\end{itemize}

This process iterates until the Blind Solver Agent's success rate draws close to a natural random chance or a predefined iteration budget is reached.
Therefore, multimodal comprehension ability is considered the primary driver for answering the refined questions correctly.
Detailed algorithmic pseudocode for this loop is provided on our website~\cite{ourwebsite}. 
Unless explicitly stated otherwise, we employ Gemini 2.5 Pro~\cite{geminiteam2025gemini25} for all MLLM-based components of the benchmark construction pipeline, including hierarchical annotation, question--answer generation, and the Solver--Critic refinement loop.

\subsection{Quality Control}
\label{sec:method:quality}

To ensure benchmark reliability, we incorporate human validation at two critical stages of the {\ourbench} construction pipeline: following the initial question-answer pair generation and after the iterative debiasing refinement stage. We formally recruited \textit{40} native Japanese experts for the initial validation and \textit{28} experts for the second-stage evaluation. The annotators verified that each QA item was grounded in the source video, culturally faithful, and appropriate for multiple-choice evaluation. In addition to improving benchmark quality through manual correction and filtering, this two-stage validation process provides an empirical assessment of the effectiveness of the automatic generation and refinement pipeline.


\textbf{Initial QA verification.} 
Across a two-week annotation phase, a panel of 40 annotators verified that every question was answerable from the video and provided a single correct answer. They directly revised ambiguous questions, overlapping options, missing correct answers, and incorrect labels.
Items that could not be reliably corrected were removed. 
Among 1,500 candidate questions, 178 received an invalid-question flag (not answerable); 161 were repaired, and 17 were removed, resulting in 1,483 candidates progressing for refinement. 
In addition, annotators authored a correct answer for 107 items and corrected the assigned label for 177 items.

\textbf{Post-refinement verification.} 
Because shortcut-triggered revision (Step 2-2) could alter the question stem, correct answer, or distractors, a second cohort of 28 annotators evaluated each refined item against its original version and the source video to ensure overall data quality. Over a two-week review stage, they confirmed question validity, verified answer correctness and specificity, and ensured distractors remained plausible, distinct, and unambiguously incorrect.
Of 1,483 refined items, 949 were accepted without a change, 532 were revised, and two were removed. The overlapping revisions included 436 answer corrections, 108 distractor edits, and 35 question rewrites. 
Following this verification, the final benchmark contains 1,481 items, and the corresponding statistics are reported in Table~\ref{tab:naru_categories}.

\input{table/dataset_statistics}

\section{Evaluation}
\label{sec:evaluation}

We evaluate both open-source and closed-source MLLMs on {\ourbench} to characterize their performance and the specific conditions under which long-form, culturally grounded understanding succeeds or breaks down. 
We begin with a full-benchmark multiple-choice evaluation to provide a controlled comparison across models and expose performance variances across the nine capability categories defined in Sec.~\ref{sec:method:taxonomy}. Because long-video understanding depends heavily on a model's temporal capacity, we systematically vary the number of sampled frames to study whether additional visual context leads to superior narrative and cultural comprehension. Subsequently, we notice that multiple-choice options can inadvertently leak contextual hints; therefore, we transform {\ourbench} into an open-ended format to reassess the models. In this setting, models are prompted to produce free-form answers that we score for factual coverage and grounding quality.

\subsection{Experimental Setup}
\label{sec:evaluation:setup}

\textbf{Models.} We evaluate a diverse set of proprietary and open-source MLLMs equipped with long-video understanding capabilities. The proprietary models comprise Gemini-3-Flash~\cite{google2025gemini3flash}, Gemini-3-Pro~\cite{google2025gemini3pro}, and Gemini-2.5-Flash~\cite{geminiteam2025gemini25}, which we access through their native video-input APIs. The open-source models comprise Qwen3.5-9B~\cite{qwen2026qwen35}, Qwen3-VL-8B~\cite{bai2025qwen3vl}, Qwen2.5-VL-7B~\cite{bai2025qwen2}, MiniCPM-o-2.6~\cite{openbmb2025minicpmo}, and InternVL3.5~\cite{wang2025internvl3}.

\textbf{Inference Setting.} For MCQ evaluation, each model answers the complete set of four-choice questions in {\ourbench}. By default, we adopt the standard input configuration recommended for each model family. For the Gemini models, we sample videos uniformly at \texttt{fps=0.25}, ensuring the native API can process the longest videos in our benchmark while maintaining consistent temporal coverage (i.e., 3600 Frames for 4-hour videos). 
For the open-source models, we apply the maximum uniform sparse-frame sampling permitted by each model's respective context window budget. 
We report overall accuracy as well as fine-grained performance across the nine capability categories: four narrative (N.1--N.4) and five cultural (C.1--C.5). For the remainder of our diagnostic experiments, we applied stratified sampling across categories to obtain 500 questions from the complete 1,481-question {\ourbench} benchmark to maintain manageable computational overhead. We use this fixed subset for all models in both the frame-budget sweep (Sec.~\ref{sec:evaluation:frame}) and the open-ended generation evaluation (Sec.~\ref{sec:evaluation:openended}).

\subsection{Overall Performance}

\input{table/rq1}

Table~\ref{tab:rq1_main} reports the primary multiple-choice results across the full benchmark. 
Overall performance follows a distinct tiering: proprietary models lead significantly, with Gemini-3-Flash achieving the highest accuracy at 76.2\%, followed by Gemini-3-Pro (70.0\%) and Gemini-2.5-Flash (51.4\%). In contrast, open-source models fall into a lower performance regime (29.6--39.8\%), led by Qwen3.5-9B.

A deeper look at category-level performance reveals a clear capability-dependent split between narrative and cultural reasoning. While Gemini models achieve approximately 11 percentage points higher accuracy on narrative tasks than on cultural ones, open-source models show virtually no difference between the two dimensions. Within narrative categories, tracking sequential structure (N.2) is universally the easiest task. Crucially, however, the primary performance bottleneck shifts as model capability increases: weaker models fail primarily at tracking low-level entity continuity (e.g., N.1 Character/Entity Evolution at 20.0\% for InternVL3.5 and 23.2\% for MiniCPM-o-2.6), whereas stronger models struggle most with high-level abstraction (N.4 Idea/Thematic Development). Thus, the dominant narrative difficulty shifts across the evaluated models, from maintaining entity continuity in most open-source models to synthesizing higher-level thematic development in the Gemini family.

The cultural domain presents an equally nuanced picture. Although open-source models perform best on C.2 (Kuuki wo Yomu), pragmatic reasoning remains challenging across all evaluated systems. In particular, C.3 (Subtext Interpretation) proves exceptionally difficult even for frontier models: Gemini-3-Flash drops to 57.4\% on C.3, despite its strong 68.2\% cultural average, making this the sole category where Gemini-3-Pro yields higher accuracy. Furthermore, several open-source models exhibit severe failure cases, with performance on narrative tracking (N.1) and cultural tasks (C.1, C.4) dipping below the 25\% random-guessing baseline. Ultimately, these findings point to two distinct mechanisms underlying performance gains: long-range narrative tracking scales directly with context length and general reasoning, whereas cultural inference is more plausibly bottlenecked by the richness and coverage of relevant knowledge in the pre-training data.

\subsection{Effect of Temporal Evidence}
\label{sec:evaluation:frame}

Long-form video understanding depends not only on model capacity but also on how densely the video is represented within the input context window.
To isolate this effect, we evaluate all models on the same 500-question subset while varying the number of sampled frames $f \in \{8,16,32,64,128\}$. 
This controlled sweep measures whether broader temporal coverage helps models recover the distributed evidence required by {\ourbench}.

\begin{figure}[t]
    \centering
    \includegraphics[width=0.85\linewidth]{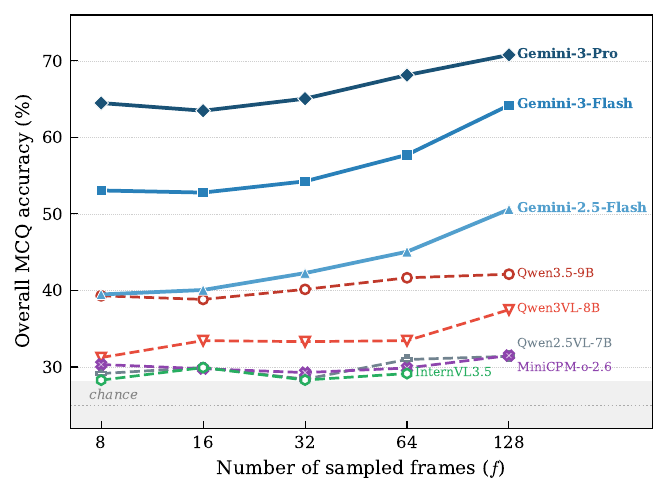}
    \caption{Overall multiple-choice accuracy as the number of sampled frames increases from 8 to 128; the dotted gray line denotes the 25\% chance level.}
    \label{fig:rq2_frame_overview}
    \vspace{-10pt}
\end{figure}

\begin{figure}[t]
    \centering
    \includegraphics[width=0.85\linewidth]{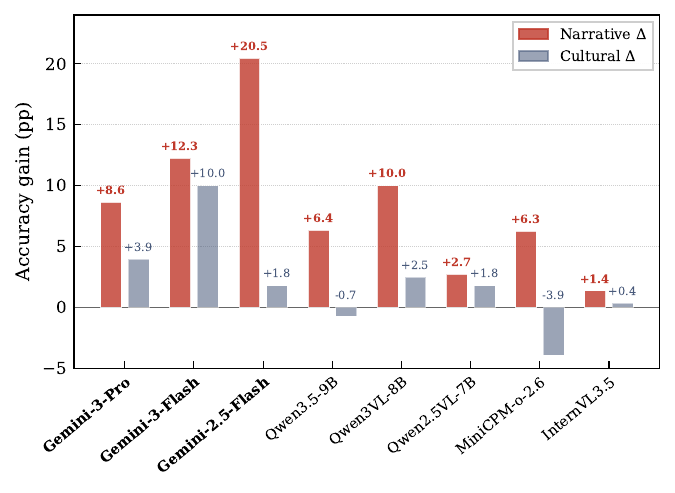} 
    \caption{Changes in narrative and cultural accuracy between 8 and 128 frames, reported in percentage points (pp).}
    \label{fig:rq2_frame_gain}
    \vspace{-10pt}
\end{figure}

Fig.~\ref{fig:rq2_frame_overview} shows the difference in how model families exploit additional temporal evidence. 
All three Gemini models improve substantially as the frame budget increases: Gemini-3-Pro rises from approximately 64\% with 8 frames to 71\% with 128 frames, Gemini-3-Flash from 53\% to 64\%, and Gemini-2.5-Flash from 39\% to 51\%. On the other hands, the open-weight models improve less consistently. Qwen3VL-8B exhibits the largest gain in this group, from about 31\% to 38\%, whereas Qwen2.5VL-7B, MiniCPM-o-2.6, and InternVL3.5 remain close to 30\% even at 128 frames. Overall, improvements are larger and more consistent among the Gemini models. One possible explanation is a difference in \emph{effective video context}: the amount of temporally distributed evidence a model can integrate, rather than its nominal input capacity. Gemini's end-to-end video pre-training and token-efficient visual encoders may help it convert additional frames into usable evidence~\cite{geminiteam2025gemini25}, whereas support for 64--128 frames in open-weight models~\cite{bai2025qwen2,openbmb2025minicpmo,wang2025internvl3} does not necessarily imply a comparable ability to select and integrate evidence across the sequence.

\input{table/rq3}

Beyond overall accuracy, the results shown in Fig.~\ref{fig:rq2_frame_gain} highlight a distinct contrast between task domains: increasing the frame budget impacts narrative understanding far more than cultural understanding. For every model, narrative performance gains consistently outpace cultural gains. Narrative accuracy rises by 1.4 to 20.5 percentage points, whereas cultural accuracy fluctuates between a 3.9-point decline and a 10.0-point gain. This disparity indicates that narrative errors are largely caused by missing dispersed events, which additional frame sampling directly resolves. Cultural interpretation, however, depends less on visual frequency and more on underlying pragmatic reasoning and domain knowledge.

Finally, a notable pattern emerges between Gemini-3-Pro and Gemini-3-Flash. Pro leads across all controlled frame budgets, though its margin over Flash shrinks from ~11 points at 8 frames to 7 points at 128 frames. However, in the full-benchmark setting at 0.25 FPS (Table~\ref{tab:rq1_main}), Flash leads Pro (76.2\% vs. 70.0\%). This reversal suggests distinct regime-dependent strengths: Pro exhibits superior low-information reasoning under strict frame constraints, whereas Flash exhibits steeper scaling dynamics, capitalizing on high-frequency temporal inputs to yield larger performance gains as context grows denser.

\subsection{Open-Ended Evaluation}
\label{sec:evaluation:openended}

MCQ evaluation provides a comparison across models, but a correct prediction may arise from recognizing the most plausible option or eliminating distractors without reconstructing the answer itself (text-only shortcut). We therefore introduce a complementary open-ended evaluation that removes all answer choices and requires each model to generate the requested information directly. 

We convert the standardized 500-question diagnostic subset into free-form questions and evaluate the same model families used in the multiple-choice experiments. Because correct responses can paraphrase the reference answer, lexical overlap and exact match are inadequate measures of correctness. We instead use GPT-5.5 as an automated judge. 
Following the idea of FActScore~\cite{min2023factscore}, for each question, the judge decomposes the reference answer into self-contained atomic facts and determines whether the generated response covers each fact. We use atomic-fact recall as the primary metric, which is defined as the fraction of reference facts covered by the response. 
 
To ensure the quality of the automated judge, we compare its results with human judgments.
That is, three annotators independently inspect 50 judged responses (sampled from Gemini-3-Flash answers) and indicate whether they agree with the judges' decisions. Across the resulting 150 verdicts, the annotators accept the judge's scores in 90.0\% of cases, with individual acceptance rates ranging from 88.0\% to 92.0\%. Majority voting accepts 48 of the 50 judgments (96.0\%), mean pairwise agreement among annotators is 82.7\%, and no judgment is rejected by all three annotators. These results support using the automated judge for aggregate analysis.

\textbf{Result.} Removing the choices preserves the performance hierarchy observed in the multiple-choice evaluation: the Gemini family maintains a substantial advantage, achieving 0.66--0.78 atomic-fact recall compared to 0.21--0.56 across open-source models (Table~\ref{tab:rq3_openqa}). However, we observe that open-ended evaluation yields a distinctly different capability profile than multiple-choice evaluation. The most striking cross-format discrepancy appears in N.2 (Sequential/Topical Flow). While N.2 is the strongest narrative subcategory across all models in the multiple-choice setting (Table~\ref{tab:rq1_main}), it degrades into the weakest narrative category for seven of the eight models under open-ended generation. A plausible explanation is that multiple-choice options serve as structural scaffolds for temporal organization, enabling models to recognize pre-sequenced event chains. Without these candidate options, models must independently retrieve, synthesize, and chronologically reconstruct events distributed across the video—making sequential flow substantially more challenging in open-ended settings.

A second notable shift is that the relative difficulty between narrative and cultural understanding reverses across formats. In the multiple-choice evaluation, six of the eight models achieve higher accuracy on narrative questions than on cultural ones. In contrast, every model achieves higher atomic-fact recall on cultural questions under open-ended evaluation. One factor that makes this reversal visible is the graded atomic-fact metric: multiple-choice evaluation applies a binary correctness criterion, whereas atomic-fact recall grants partial credit whenever a response recovers a subset of ground-truth facts. Across all models, 78.9\% of cultural responses recover at least one reference fact, compared to 66.3\% of narrative responses. Because cultural prompts often outline the underlying interaction before asking for its implicit significance, this contextual grounding helps models generate partially correct observations even when missing the full interpretation. Conversely, narrative questions frequently demand the unprompted reconstruction of distributed events, leading to a higher rate of zero-recall responses.


\section{Conclusion}

This paper introduced {\ourbench}, a benchmark of 1,481 questions grounded in 155 Japanese long-form videos totalling 146.8 hours and spanning four narrative and five cultural dimensions. {\ourbench} combines a hierarchical annotation-to-QA pipeline with iterative shortcut removal and two-stage verification by 68 native Japanese annotators to construct high-quality questions at this temporal scale. Evaluation across various model configurations shows that both performance and behavior vary substantially across model families, including how models respond to denser frame sampling and how they behave under open-ended settings. 
Crucially, our results show that current open-source models lag significantly behind top commercial models in both narrative reasoning and culturally grounded understanding. We hope {\ourbench} serves as a foundational catalyst for developing next-generation MLLMs capable of genuine, high-context video understanding.

\bibliographystyle{IEEEtran} 
\bibliography{ref}           

\end{document}

%% file: micro.tex
\usepackage{enumitem}
\usepackage{algorithm}
\usepackage{algpseudocode}
\usepackage{amsmath}
\usepackage{xspace}
\usepackage{listings}
\usepackage{xcolor}
\usepackage{subcaption}
\usepackage{graphicx}
\usepackage{url}
\usepackage{booktabs}
\usepackage{tabularx}
\usepackage{multirow}
\usepackage{array}
\usepackage{xeCJK}
\setCJKsansfont{HaranoAjiGothic-Medium.otf}

\def\ourbench{\textsc{NARU}}

\makeatletter
\DeclareRobustCommand\onedot{\futurelet\@let@token\@onedot}
\def\@onedot{\ifx\@let@token.\else.\null\fi\xspace}
\def\eg{\emph{e.g}\onedot}

\def\wrt{w.r.t\onedot}

%% file: table/dataset_statistics.tex
\begin{table*}[t]
\centering
\footnotesize
\setlength{\tabcolsep}{5pt}
\renewcommand{\arraystretch}{0.8}
\begin{tabular}{>{\centering\arraybackslash}p{0.08\linewidth}p{0.33\linewidth}p{0.3\linewidth}cr}
\toprule
\textbf{Level} & \textbf{Task} & \textbf{Type of Evidence}  & \textbf{Code} & \textbf{\#} \\
\midrule

\multirow{4}{*}{\shortstack[c]{\textbf{Narrative}\\(N, 745)}}
& \textsc{Character/Entity Evolution}
& A character/entity across segments & N.1 & 185 \\

& \textsc{Sequential/Topical Flow}
& Events or topics over time
& N.2 & 187 \\

& \textsc{Plot/Conflict Progression}
& A causal thread or conflict
& N.3 & 186 \\

& \textsc{Idea/Thematic Development}
& Motifs, claims, or narrative cues
& N.4 & 187 \\

\midrule

\multirow{5}{*}{\shortstack[c]{\textbf{Cultural}\\(C, 736)}}
& \textsc{Aizuchi} (Conversational Mechanics)
& Backchannels and response timing
& C.1 & 143 \\

& \textsc{Kuuki wo Yomu} (Situational Awareness)
& Social atmosphere or implicit norms
& C.2 & 147 \\

& \textsc{Subtext Interpretation}
& Surface utterance plus context
& C.3 & 148\\

& \textsc{Cultural Context Recognition}
& Culturally specific references
& C.4 & 149 \\

& \textsc{Sentiment Analysis}
& Verbal, visual, and social cues
& C.5 & 149 \\

\bottomrule
\end{tabular}
\caption{Statistics of {\ourbench}.}
\label{tab:naru_categories}
\vspace{-8pt}
\end{table*}

%% file: table/rq1.tex
\begin{table*}[t]
\centering
\small
\setlength{\tabcolsep}{2.4pt}
\begin{tabular}{l c c c c c c c c c c c c c}
\toprule
Model & Sampling Rate & N.1 & N.2 & N.3 & N.4 & Narr. Avg & C.1 & C.2 & C.3 & C.4 & C.5 & Cult. Avg & Overall \\
\midrule
Gemini-3-Flash & 0.25 FPS & \textbf{83.8} & \textbf{90.9} & \textbf{83.9} & \textbf{78.1} & \textbf{84.2} & \textbf{71.3} & \textbf{69.4} & \underline{57.4} & \textbf{73.2} & \textbf{69.8} & \textbf{68.2} & \textbf{76.2} \\
Gemini-3-Pro & 0.25 FPS & \underline{74.6} & \underline{78.6} & \underline{76.3} & \underline{66.8} & \underline{74.1} & \underline{60.1} & \underline{68.7} & \textbf{64.2} & \underline{69.8} & \underline{67.1} & \underline{66.0} & \underline{70.0} \\
Gemini-2.5-Flash & 0.25 FPS & 51.3 & 63.1 & 58.1 & 50.8 & 55.8 & 37.8 & 49.7 & 48.6 & 50.3 & 48.3 & 46.9 & 51.4 \\
\midrule
Qwen3.5-9B & 128 Frames & 34.6 & 48.7 & 34.9 & 34.2 & 38.1 & 41.3 & 49.0 & 34.5 & 40.9 & 41.6 & 41.4 & 39.8 \\
Qwen3VL-8B & 0.25 FPS & 35.7 & 46.5 & 39.8 & 35.8 & 39.5 & 32.2 & 40.1 & 33.1 & 38.9 & 32.9 & 35.4 & 37.4 \\
Qwen2.5VL-7B & 128 Frames & 25.4 & 42.2 & 30.6 & 26.2 & 31.1 & 22.4 & 40.8 & 25.7 & 24.2 & 28.2 & 28.2 & 29.7 \\
MiniCPM-o-2.6 & 128 Frames & 23.2 & 41.2 & 27.4 & 31.6 & 30.8 & 25.9 & 37.4 & 27.7 & 22.1 & 28.2 & 28.3 & 29.6 \\
InternVL3.5 & 64 Frames & 20.0 & 36.4 & 25.8 & 31.0 & 28.3 & 32.2 & 41.5 & 31.8 & 35.6 & 32.2 & 34.6 & 31.5 \\
\bottomrule
\end{tabular}
\caption{MCQ accuracy results on NARU. The \textbf{best} and the \underline{2nd best} results are marked.
}
\label{tab:rq1_main}
\vspace{-10pt}
\end{table*}

%% file: table/rq3.tex
\begin{table}[!t]
\centering
\vspace{2pt}
\setlength{\tabcolsep}{1.5pt}
\renewcommand{\arraystretch}{1.05}
\resizebox{\columnwidth}{!}{%
\begin{tabular}{lcccccccccc}
\toprule
Model & N.1 & N.2 & N.3 & N.4 & C.1 & C.2 & C.3 & C.4 & C.5 & Avg \\
\midrule
Gemini-3-Flash & \textbf{0.78} & \textbf{0.66} & \textbf{0.72} & \textbf{0.75} & \textbf{0.69} & \textbf{0.87} & \textbf{0.93} & \textbf{0.85} & \textbf{0.80} & \textbf{0.78} \\
Gemini-3-Pro & \underline{0.77} & \underline{0.61} & \underline{0.71} & \underline{0.69} & \underline{0.65} & \underline{0.87} & \underline{0.88} & \underline{0.80} & 0.72 & \underline{0.75} \\
Gemini-2.5-Flash & 0.62 & 0.50 & 0.58 & 0.65 & 0.56 & 0.77 & 0.84 & 0.68 & \underline{0.73} & 0.66 \\
Qwen3.5-9B & 0.54 & 0.49 & 0.47 & 0.50 & 0.52 & 0.66 & 0.65 & 0.60 & 0.59 & 0.56 \\
Qwen3-VL-8B & 0.39 & 0.23 & 0.34 & 0.36 & 0.47 & 0.64 & 0.62 & 0.39 & 0.53 & 0.44 \\
Qwen2.5-VL-7B & 0.35 & 0.24 & 0.27 & 0.28 & 0.43 & 0.47 & 0.40 & 0.28 & 0.40 & 0.35 \\
MiniCPM-o-2.6 & 0.21 & 0.12 & 0.15 & 0.16 & 0.23 & 0.26 & 0.30 & 0.15 & 0.31 & 0.21 \\
InternVL3.5 & 0.40 & 0.14 & 0.27 & 0.35 & 0.43 & 0.49 & 0.47 & 0.39 & 0.51 & 0.38 \\
\bottomrule
\end{tabular}%
}
\caption{Open-ended correctness on NARU. Scores are FActScore recall (0--1): fraction of reference atomic facts covered by the answer.
}
\label{tab:rq3_openqa}
\vspace{-15pt}
\end{table}